\documentclass[11pt, a4paper, logo, copyright, nonumbering]{deepseek}
\usepackage[authoryear, sort&compress, round]{natbib}
\usepackage{dblfloatfix}
\usepackage{ulem}
\usepackage{caption}
\usepackage{dramatist}
\usepackage{xspace}
\usepackage{pifont}
\usepackage{multirow}
\usepackage{adjustbox}
\usepackage{tcolorbox}
\usepackage{xltabular}
\usepackage{longtable}
\usepackage{hyperref}
\usepackage{makecell}
\usepackage{algorithm}
\usepackage[noend]{algpseudocode}

\usepackage{amsfonts}
\usepackage{amsmath}
\usepackage{amssymb}
\usepackage{lineno}
\usepackage{wrapfig}

\usepackage[bottom]{footmisc}

\usepackage{subcaption}
\usepackage{setspace}

\usepackage{titlesec}
\usepackage{enumitem}

\makeatletter
\def\@BTrule[#1]{%
  \ifx\longtable\undefined
    \let\@BTswitch\@BTnormal
  \else\ifx\hline\LT@hline
    \nobreak
    \let\@BTswitch\@BLTrule
  \else
     \let\@BTswitch\@BTnormal
  \fi\fi
  \global\@thisrulewidth=#1\relax
  \ifnum\@thisruleclass=\tw@\vskip\@aboverulesep\else
  \ifnum\@lastruleclass=\z@\vskip\@aboverulesep\else
  \ifnum\@lastruleclass=\@ne\vskip\doublerulesep\fi\fi\fi
  \@BTswitch}
\makeatother

\addto\extrasenglish{
}

 {\begin{list}{}%
         {\setlength{\leftmargin}{#1}}%
         \item[]%
 }
 {\end{list}}

\reportnumber{001} 

\renewcommand{\today}{}

\title{\centering 60 Data Points are Sufficient to Fine-Tune LLMs for Question-Answering}
\author[*]{
\bf{
Junjie Ye$^{1*}$, Yuming Yang$^{1*}$, Qi Zhang$^{1\dag}$, Tao Gui$^{2}$,} \\
\bf{Xuanjing Huang$^{1}$, Peng Wang$^{3}$, Zhongchao Shi$^{3}$, Jianping Fan$^{3}$}
\\
{$^1$ School of Computer Science, Fudan University} \\
  {$^2$ Institute of Modern Languages and Linguistics, Fudan University} \\
  {$^{3}$ Lenovo Research, Beijing, China}\\
\texttt{jjye23@m.fudan.edu.cn, \{qz, tgui\}@fudan.edu.cn} 
}
\correspondingauthor{\, $^*$ Equal Contribution. \\\, $^\dag$ Corresponding Author.}

\begin{abstract}

Large language models (LLMs) encode extensive world knowledge through pre-training on massive datasets, which can then be fine-tuned for the question-answering (QA) task. However, effective strategies for fine-tuning LLMs for the QA task remain largely unexplored. To address this gap, we categorize supervised fine-tuning (SFT) data based on the extent of knowledge memorized by the pretrained LLMs and conduct a series of empirical analyses. Our experiments, involving four LLMs from three different model families, focus on three key factors: \textit{the amount of data required for SFT}, \textit{the impact of different SFT datasets on model performance}, and \textit{how data requirements vary across LLMs}. The results show that as few as 60 data points during the SFT stage can activate the knowledge encoded during pre-training, enabling LLMs to perform the QA task. Additionally, SFT with data of varying memory levels has a significant impact on LLM performance, with the optimal dataset differing based on the specific model being fine-tuned. Future research will delve deeper into the mechanisms underlying these phenomena.

\end{abstract}

\begin{document}
\maketitle

\section{Introduction}

Large Language Models (LLMs), such as the GPT~\citep{GPT-3, InstructGPT, Codex, GPT-4}, LLaMA~\citep{LLaMA-1, CodeLLaMA, LLaMA-2, LLaMA-3}, and Qwen~\citep{Qwen-1, Qwen-2} series, are pretrained on diverse corpora covering a wide range of genres and world knowledge. This knowledge is encoded within the model's parameters~\citep{analy-ye, base-1, base-2, base-3, base-4} and can be applied to the question-answering (QA) task through supervised fine-tuning (SFT)~\citep{QA-1, QA-2, QA-3}.

As research advances, there is growing interest in optimizing fine-tuning strategies for LLMs in the QA task. For instance,~\cite{sft-renmin} collected a dataset of multiple-choice questions, segmented the data based on the accuracy of pretrained LLM responses, and conducted a instruction fine-tuning study. Their findings suggest that effective SFT requires maintaining consistent knowledge within model parameters before and after fine-tuning. Similarly, using a Wikipedia-based corpus,~\cite{sft-google} segmented data based on the accuracy of pretrained models with different hyperparameters and found that training on poorly memorized knowledge from pretrained LLMs significantly increased hallucinations.

However, these studies fall short in identifying effective fine-tuning strategies for LLMs in the QA task. On one hand, they segment the training data based on the accuracy of LLM responses in few-shot scenarios, which are highly sensitive to specific in-context examples and may introduce bias into the results~\citep{in-context-1, in-context-2}. On the other hand, the granularity of data segmentation is relatively coarse, typically limited to three categories (e.g., low, medium, and high), restricting detailed analysis and hindering the ability to draw more precise conclusions.

To explore this issue in depth, we develop a robust multi-template complementation mechanism (Section~\ref{sec:complementation}) to evaluate how well pretrained LLMs memorize different types of knowledge. We then conduct an empirical analysis of four LLMs from three families to address three key questions.

\textbf{Q1: How much data is needed in the SFT stage to enable LLMs to perform the QA task?}
We collect data from Wikipedia on 12 location-related topics to create the training and test sets, as well as data on 12 unrelated topics to build the out-of-domain test set (Section~\ref{sec:dataset}). By varying the amount of training data, we find that only 60 data points are needed in the SFT stage for LLMs to efficiently perform the QA task and demonstrate strong generalization ability. Increasing the training data does not yield significant gains and may even harm model performance (Section~\ref{sec:amount}). \textit{We speculate that this is because SFT activates and refines knowledge already encoded during pre-training, requiring only minimal parameter tuning to optimize the process.}

\textbf{Q2: How do different SFT datasets affect LLM performance on the QA task?}
We investigate this by categorizing training and test data into five memory levels using the multi-template complementation mechanism. Our findings reveal that fine-tuning with data of varying memory levels results in significant differences in knowledge activation. Specifically, while LLMs consistently provide more accurate answers to knowledge that is well-remembered from pre-training, using data that the model barely memorized to do SFT severely impairs its activation of high-memory-level knowledge (Section~\ref{sec:memory}). \textit{This underscores the importance of careful data selection in SFT and demonstrates how different datasets can profoundly impact the ability of LLMs to perform the QA task.}

\textbf{Q3: How do data requirements for the SFT stage vary across LLMs?}
We conduct a comparative analysis of the knowledge memory levels of different LLMs (Section~\ref{sec:distribution}) and fine-tune them using the same data, observing significant differences in their performance on the QA task (Section~\ref{sec:compare}). \textit{This suggests that variations in the pre-training corpora of LLMs lead to corresponding differences in their data requirements during the SFT stage, offering new insights into the optimal composition of training data for different models.}

In summary, our contributions are as follows:
\begin{itemize} 
    \item We design a multi-template complementation mechanism that reliably assesses the extent to which pretrained LLMs memorize different types of knowledge. 
    \item We conduct an extensive empirical analysis of four LLMs from three different families to address three key questions regarding fine-tuning LLMs for the QA task.
    \item We identify intrinsic differences in fine-tuning different LLMs for the QA task, offering new insights to guide the development of more effective fine-tuning strategies. 
    \item We plan to further explore the underlying mechanisms of fine-tuning LLMs for the QA task, providing deeper explanations for these findings.
\end{itemize}

\section{Experimental Setup}

We perform a comprehensive empirical analysis to offer insights into fine-tuning LLMs for the QA task. The corresponding experimental setup is detailed in this section.

\subsection{Multi-Template Complementation Mechanism}
\label{sec:complementation}

\begin{figure}[!t]
    \centering
        \includegraphics[width=0.99\linewidth]{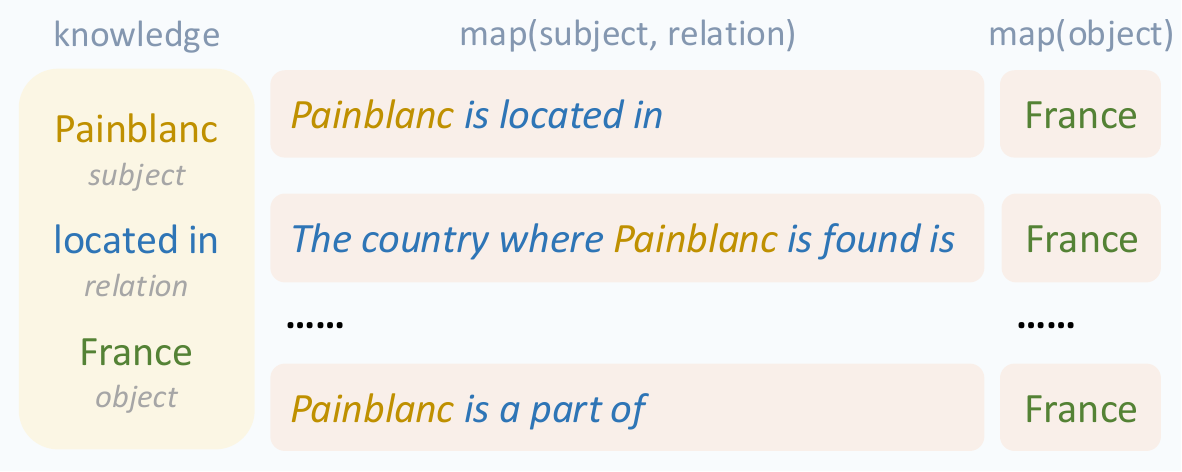}
    \caption{An example for the multi-template complementation mechanism.}
    \label{fig:example}
\end{figure}

Given a pre-trained LLM $M_{base}$ and a knowledge dataset $K$, we aim to explore how fine-tuning $M_{base}$ with a subset of $K$, denoted as $K_i \subseteq K$, impacts its performance on the QA task. Previous studies have categorized data based on the accuracy of $M_{base}$'s responses in few-shot scenarios, which can be biased by the specific examples used. To address this, we propose a memory discrimination scheme utilizing a more robust multi-template complementary mechanism.

\begin{wraptable}{!t}{0.5\linewidth}
    \centering
    \caption{Details of the dataset used. `\# Topics' and `\# Number' represent the number of topics and the amount of data in the corresponding dataset, respectively.}
    {
    \begin{tabular}{l|cc}
    \toprule
       \textbf{Data}  & \textbf{\# Topics} & \textbf{\# Number} \\ \midrule
        $D_{train}$ & 12 & 92435 \\
        $D_{test}$ & 12 & 11546 \\
        $D_{test-ood}$ & 12 & 9660 \\ \bottomrule
    \end{tabular}
    }
    \label{tab:data}
\end{wraptable}

Illustrated as Figure~\ref{fig:example}, consider $k \in K$ as an element in $K$ represented by a triple $(subject, relation, object)$, such as $(Painblanc, located in, France)$. Given a sentence $x = map(subject, relation)$ that maps the subject and relation (e.g., `\textit{Painblanc is located in }'), if $M_{base}$ can predict $y = map(object)$ by mapping the object (e.g., `\textit{France}'), such that $y \subseteq M_{base}(x)$, we consider that $M_{base}$ has memorized knowledge $k$.
Since $M_{base}$ is a probabilistic model whose output is influenced by different mapping templates and sampling probabilities, we design $N_{map} = 21$ different mappings for each piece of knowledge $k$. With the temperature set to 0.7, the model generates $N_{sample} = 10$ outputs for each mapping, and we calculate the degree to which the LLM memorizes $k$ as 
\begin{equation}
    R_k^M = \frac{\sum_{i=1}^{N_{map}} \sum_{j=1}^{N_{sample}} \mathbb{I}(y_i \subseteq M_{base}^j(x_i))}{N_{map} \times N_{sample}}
\end{equation}
where $x_i$ and $y_i$ are the results from the $i$th mapping, $M_{base}^j$ represents the $j$th sample, and $\mathbb{I}(\cdot)$ is the indicator function.

This method leverages the LLM's text-completion capabilities while addressing the bias from contextual examples in previous studies. By using multiple templates and repeated sampling, it reduces the impact of randomness in probabilistic model sampling, effectively measuring the model's memory of knowledge.

\subsection{Dataset}
\label{sec:dataset}

For our study, we use ENTITYQUESTIONS~\citep{EntityQuestions}, a QA dataset containing knowledge from Wikipedia on 24 different topics. We select the original training sets from 12 location-related topics as our training data $D_{train}$, their corresponding test sets as our test sets $D_{test}$, and the test sets from the remaining 12 topics as out-of-domain test sets $D_{test-ood}$. Details of the data are listed in Table~\ref{tab:data}. Detailed topics and corresponding mapping templates can be found in Appendix~\ref{sec:detail_data}.

\subsection{Models}

To ensure generalizable results, we analyze four LLMs from three different families, described as follows.

\begin{itemize}
    \item \textbf{LLaMA-2 Family.} The LLaMA-2 family~\citep{LLaMA-2} consists of open-source LLMs developed by Meta, pre-trained on over 2 trillion tokens, offering extensive world knowledge and strong semantic representation. In this paper, we select \textit{LLaMA-2-7B} and \textit{LLaMA-2-13B} for our analysis.
    \item \textbf{LLaMA-3 Family.} The LLaMA-3 family~\citep{LLaMA-3} represents the latest and most advanced open-source LLMs developed by Meta, available in both 8B and 70B parameter sizes. In this paper, we select \textit{LLaMA-3-8B} for our analysis.
    \item \textbf{Qwen-2 Family.} The Qwen-2 family~\citep{Qwen-2}, developed by Alibaba, consists of generalized LLMs trained on data in 29 languages, with parameter sizes ranging from 0.5B to 72B. In this paper, we select \textit{Qwen-2-7B} for our analysis.
\end{itemize}

\subsection{Metrics}
\label{sec:metrics}

Given a pre-traind LLM $M_{base}$, to provide a detailed analysis of its performance on the QA task after SFT, we apply the mechanism defined in Section~\ref{sec:complementation} and divide the test set based on memorization levels:

\begin{equation}
\left\{
\begin{aligned}
D_{test-0} &= \{k \in D_{test} | R_k^M = 0\}, \\
D_{test-1} &= \{k \in D_{test} | R_k^M \in (0, 0.25]\}, \\
D_{test-2} &= \{k \in D_{test} | R_k^M \in (0.25, 0.5]\}, \\
D_{test-3} &= \{k \in D_{test} | R_k^M \in (0.5, 0.75]\}, \\
D_{test-4} &= \{k \in D_{test} | R_k^M \in (0.75, 1]\}
\end{aligned}
\right.
\end{equation}

We calculate the proportion of $y \subseteq M_{sft}(x)$ under the test mapping template to determine the accuracy for each sub-test set, $ACC_{test-i}$ ($i = 0, 1, 2, 3, 4$), where $M_{sft}$ represents for the LLM after fine-tuning. The overall performance on the test set is then obtained by averaging the accuracies:

\begin{equation}
    ACC_{test} = AVG_{i=0}^4(ACC_{test-i})
\end{equation}

Furthermore, we apply the same method to the train set $D_{train}$ and the out-of-domain test set $D_{test-ood}$, yielding $D_{train-i}$, $D_{test-ood-i}$, $ACC_{test-ood}$, and other metrics, where $i = 0, 1, 2, 3, 4$.

\subsection{Implementation Details}

\begin{figure}[!t]
    \centering
        \includegraphics[width=\linewidth]{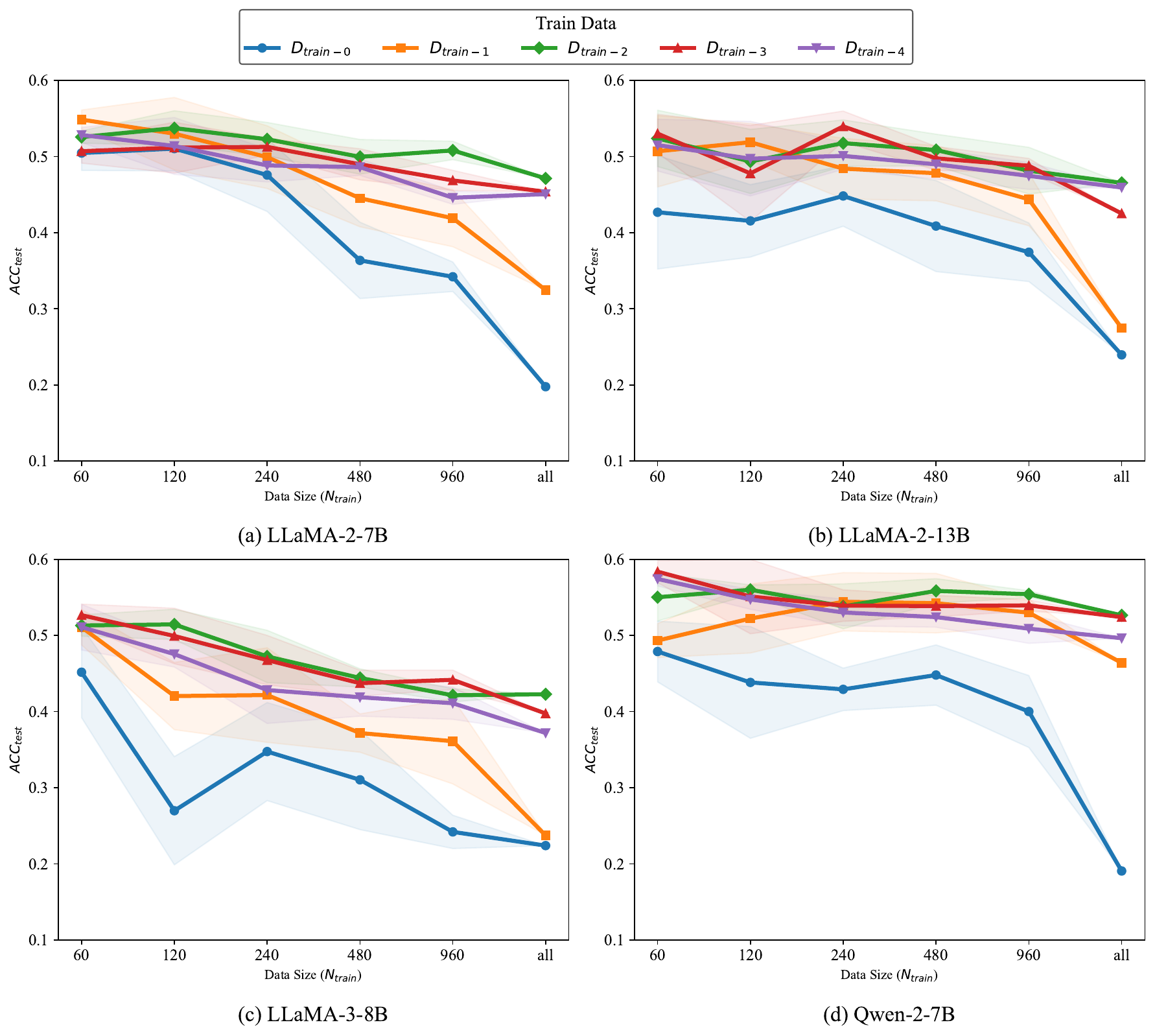}
    \caption{Performance (in-domain) of LLMs trained using different amounts of data. Each line in the plot represents training with data from a specific memory level.}
    \label{fig:id_size}
\end{figure}

Our experiments are divided into three main stages: memory level differentiation, SFT, and performance evaluation. Below are the implementation details for each stage:
\begin{itemize}
    \item \textbf{Memory Level Differentiation.} To balance output stability and diversity, we design 21 mapping templates for the data of each topic. The temperature is set to 0.7 for model sampling, with 10 repetitions for each sample. The output token length is set to 32.
    \item \textbf{SFT.} In this stage, we set the batch size to 16, and train for 1 epoch using the AdamW~\citep{Adamw} optimizer with cosine scheduling. The learning rate is set to 2e-5 for the LLaMA-2 models and 1e-5 for the others. To fully utilize model capabilities, we use the officially recommended prompt templates for each model, as listed in Appendix~\ref{sec:prompt}.
    \item \textbf{Performance Evaluation.} For sampling, we use greedy search with a maximum output length of 16, maintaining the same prompt templates as during training. To reduce bias from the selected training data, we randomly select five different sets of training data for each memory level and repeat the experiments with these sets. We then report the mean and variance of the results from these five experiments.
\end{itemize}

\section{Main Results}
\label{sec:result}

To provide a comprehensive analysis of how to effectively fine-tune LLMs for QA tasks, we examine the data volume requirements during the SFT stage (Section~\ref{sec:amount}) and the impact of fine-tuning with data at different memory levels (Section~\ref{sec:memory}).

\subsection{Data Volume Requirements during SFT}
\label{sec:amount}

\begin{figure}[!t]
    \centering
        \includegraphics[width=\linewidth]{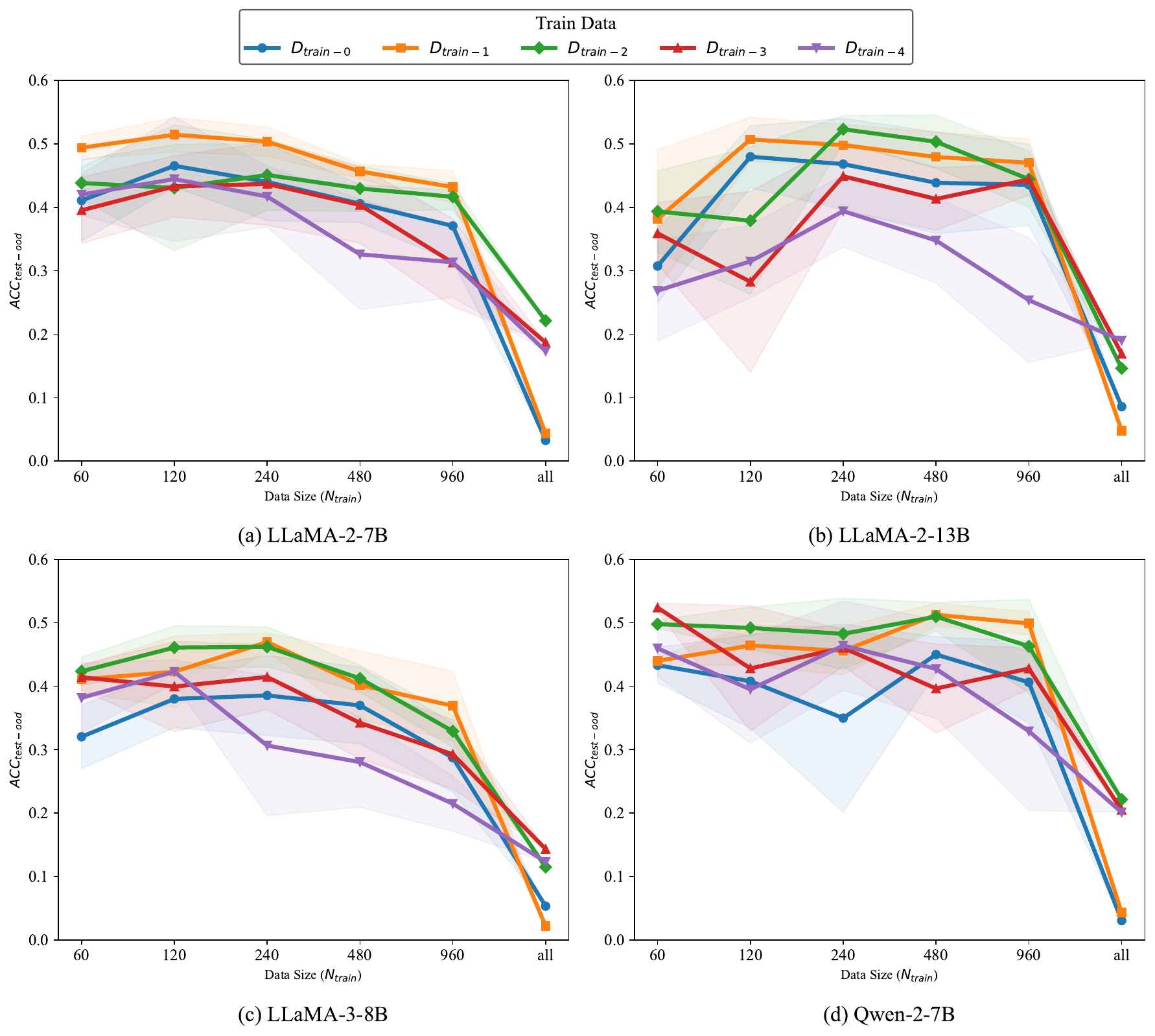}
    \caption{Performance (out-of-domain) of LLMs trained using different amounts of data. Each line in the plot represents training with data from a specific memory level.}
    \label{fig:ood_size}
\end{figure}

Focusing on Q1—how much data is needed in the SFT stage to enable LLMs to perform the QA task—we analyze each LLM \( M_{base} \) using different memory levels of the training data \( D_{train-i}^M \). We divide the training data into six volume levels, ranging from 60 samples to the full dataset, and compose the training sets by uniformly sampling from 12 topics. The results of the in-domain and out-of-domain evaluation experiments under different settings are shown in Figure~\ref{fig:id_size} and Figure~\ref{fig:ood_size}, respectively.

\paragraph{Finding 1}  
The experimental results indicate that just 60 training samples are sufficient for LLMs to efficiently perform the QA task after SFT while demonstrating strong generalization ability. The in-domain results (Figure~\ref{fig:id_size}) reveal that LLMs perform better with fewer training samples compared to using 960 or all samples, regardless of the base model or memory level. Most models peak or approach peak performance at \( N_{train} = 60 \), suggesting this amount is adequate for the in-domain task. Furthermore, the out-of-domain evaluation results (Figure~\ref{fig:ood_size}) indicate that optimal performance can still be achieved with a limited number of training samples compared to the full dataset. We speculate that this occurs because SFT activates and refines knowledge already encoded during pre-training, requiring only minimal parameter tuning to optimize the process.

\subsection{Impact of Fine-Tuning with Data at Different Memory Levels}
\label{sec:memory}

Consider Q2, i.e., how do different SFT datasets affect LLM performance on the QA task? Following the method outlined in Section~\ref{sec:metrics}, we categorize the memory levels of the training data, test data, and out-of-domain test data for each LLM. Since we established in Section~\ref{sec:amount} that fine-tuning with just 60 samples enables LLMs to perform satisfactorily on the QA task, this section focuses solely on the results under this setup. We present the mean and standard deviation of performance for different LLMs on the in-domain and out-of-domain test sets in Table~\ref{tbl:bucket_id} and Table~\ref{tbl:bucket_ood}, respectively.

\paragraph{Finding 2} Regardless of the data used for fine-tuning, LLMs consistently provide more accurate answers to knowledge that is better memorized during pre-training. Examining each row in Table~\ref{tbl:bucket_id} and Table~\ref{tbl:bucket_ood}, we observe that SFT models consistently achieves better performance on test sets with higher memory levels compared to those with lower memory levels, as indicated by the relationship \( ACC_{test-4} > ACC_{test-3} > ACC_{test-2} > ACC_{test-1} > ACC_{test-0} \).

\paragraph{Finding 3} Training with data at a specific memory level enhances the performance of LLMs on that level of knowledge. In Table~\ref{tbl:bucket_id}, we notice an interesting `diagonal phenomenon': for test data of a particular memory level, LLMs trained with data of the same memory level tend to perform the best. For example, for \( D_{test-0} \), the performance using \( D_{train-0} \) is typically the highest. This suggests that data from different memory levels may be encoded differently within the model. Therefore, selecting the appropriate dataset is crucial when aiming to enhance LLM performance across different knowledge levels. Table~\ref{tbl:bucket_ood} also demonstrates a similar phenomenon in the out-of-domain test set.

\paragraph{Finding 4} Overall, a more effective strategy is to use data with higher memory levels for SFT. Table~\ref{tbl:bucket_id} and Table~\ref{tbl:bucket_ood} show that training with \( D_{train-0} \) significantly impairs LLM performance on higher memory levels of the test data (e.g., \( D_{test-3} \), \( D_{test-4} \)), negatively affecting overall performance. In contrast, models trained with relatively high memory level data tend to achieve the best overall performance \( ACC_{test} \) because they maintain a more balanced approach across different memory levels.

\begin{table}[!t]
    \centering
\caption{Performance (in-domain) of LLMs trained using data with different memory levels on test sets with different memory levels. The best performance of each test set is \textbf{bolded}.}
    
    \begin{subtable}{\linewidth}
        \centering
        \caption{LLaMA-2-7B ($N_{train}=60$)}
        \resizebox{\linewidth}{!}{
        \begin{tabular}{l|ccccc|c}
            \toprule
            \textbf{Train Data} & {$\mathbf{ACC_{test-0}}$} & $\mathbf{ACC_{test-1}}$ & $\mathbf{ACC_{test-2}}$ & $\mathbf{ACC_{test-3}}$ & $\mathbf{ACC_{test-4}}$ & $\mathbf{ACC_{test}}$ \\ 
            \midrule
$D_{train-0}$ & \textbf{4.80} $\pm$ \scriptsize{1.11} & 34.00 $\pm$ \scriptsize{2.93} & 62.49 $\pm$ \scriptsize{4.83} & 73.36 $\pm$ \scriptsize{4.13} & 77.64 $\pm$ \scriptsize{6.02} & 50.46 $\pm$ \scriptsize{2.28} \\
$D_{train-1}$ & 4.12 $\pm$ \scriptsize{1.61} & \textbf{38.19} $\pm$ \scriptsize{4.54} & 69.49 $\pm$ \scriptsize{2.23} & 77.11 $\pm$ \scriptsize{3.43} & 85.32 $\pm$ \scriptsize{1.62} & \textbf{54.85} $\pm$ \scriptsize{1.26} \\
$D_{train-2}$ & 0.69 $\pm$ \scriptsize{0.35} & 24.57 $\pm$ \scriptsize{1.70} & \textbf{70.57} $\pm$ \scriptsize{2.23} & 82.39 $\pm$ \scriptsize{1.65} & 84.56 $\pm$ \scriptsize{0.64} & 52.56 $\pm$ \scriptsize{0.81} \\
$D_{train-3}$ & 0.37 $\pm$ \scriptsize{0.41} & 17.79 $\pm$ \scriptsize{2.33} & 67.45 $\pm$ \scriptsize{2.34} & 85.43 $\pm$ \scriptsize{2.13} & 82.45 $\pm$ \scriptsize{2.84} & 50.70 $\pm$ \scriptsize{1.57} \\
$D_{train-4}$ & 0.40 $\pm$ \scriptsize{0.28} & 19.78 $\pm$ \scriptsize{1.67} & 69.77 $\pm$ \scriptsize{1.24} & \textbf{87.36} $\pm$ \scriptsize{1.45} & \textbf{86.84} $\pm$ \scriptsize{0.91} & 52.83 $\pm$ \scriptsize{0.99} \\
            \bottomrule
        \end{tabular}}
    \end{subtable}%
    
    \vspace{6pt} 
    
    \begin{subtable}{\linewidth}
        \centering
        \caption{LLaMA-2-13B ($N_{train}=60$)}
        \resizebox{\linewidth}{!}{
        \begin{tabular}{l|ccccc|c}
            \toprule
            \textbf{Train Data} & $\mathbf{ACC_{test-0}}$ & $\mathbf{ACC_{test-1}}$ & $\mathbf{ACC_{test-2}}$ & $\mathbf{ACC_{test-3}}$ & $\mathbf{ACC_{test-4}}$ & $\mathbf{ACC_{test}}$ \\ 
            \midrule
$D_{train-0}$ & \textbf{7.62} $\pm$ \scriptsize{1.69} & 23.66 $\pm$ \scriptsize{4.29} & 53.17 $\pm$ \scriptsize{9.36} & 61.96 $\pm$ \scriptsize{11.69} & 67.00 $\pm$ \scriptsize{14.25} & 42.68 $\pm$ \scriptsize{7.45} \\
$D_{train-1}$ & 7.49 $\pm$ \scriptsize{0.54} & \textbf{35.95} $\pm$ \scriptsize{5.15} & 64.22 $\pm$ \scriptsize{5.00} & 65.98 $\pm$ \scriptsize{5.28} & 79.84 $\pm$ \scriptsize{9.43} & 50.69 $\pm$ \scriptsize{4.69} \\
$D_{train-2}$ & 0.62 $\pm$ \scriptsize{0.50} & 21.36 $\pm$ \scriptsize{2.93} & \textbf{72.80} $\pm$ \scriptsize{5.72} & 80.65 $\pm$ \scriptsize{5.45} & 86.47 $\pm$ \scriptsize{5.85} & 52.38 $\pm$ \scriptsize{3.70} \\
$D_{train-3}$ & 0.45 $\pm$ \scriptsize{0.21} & 17.97 $\pm$ \scriptsize{1.81} & 71.16 $\pm$ \scriptsize{3.63} & \textbf{86.60} $\pm$ \scriptsize{3.40} & \textbf{88.87} $\pm$ \scriptsize{4.79} & \textbf{53.01} $\pm$ \scriptsize{2.54} \\
$D_{train-4}$ & 0.24 $\pm$ \scriptsize{0.22} & 16.66 $\pm$ \scriptsize{3.14} & 68.46 $\pm$ \scriptsize{4.83} & 83.44 $\pm$ \scriptsize{4.92} & 88.69 $\pm$ \scriptsize{4.82} & 51.50 $\pm$ \scriptsize{3.40} \\

            \bottomrule
        \end{tabular}}
    \end{subtable}
    
    \vspace{6pt} 
    
    \begin{subtable}{\linewidth}
        \centering
        \caption{LLaMA-3-8B ($N_{train}=60$)}
        \resizebox{\linewidth}{!}{
        \begin{tabular}{l|ccccc|c}
            \toprule
            \textbf{Train Data} & $\mathbf{ACC_{test-0}}$ & $\mathbf{ACC_{test-1}}$ & $\mathbf{ACC_{test-2}}$ & $\mathbf{ACC_{test-3}}$ & $\mathbf{ACC_{test-4}}$ & $\mathbf{ACC_{test}}$ \\ 
            \midrule
$D_{train-0}$ & \textbf{2.72} $\pm$ \scriptsize{1.44} & 26.05 $\pm$ \scriptsize{8.77} & 53.82 $\pm$ \scriptsize{10.32} & 62.90 $\pm$ \scriptsize{8.27} & 80.36 $\pm$ \scriptsize{9.57} & 45.17 $\pm$ \scriptsize{5.96} \\
$D_{train-1}$ & 2.28 $\pm$ \scriptsize{1.23} & \textbf{33.83} $\pm$ \scriptsize{7.50} & 66.35 $\pm$ \scriptsize{3.69} & 71.97 $\pm$ \scriptsize{4.08} & 81.02 $\pm$ \scriptsize{3.91} & 51.09 $\pm$ \scriptsize{2.33} \\
$D_{train-2}$ & 0.19 $\pm$ \scriptsize{0.08} & 21.57 $\pm$ \scriptsize{1.71} & \textbf{71.89} $\pm$ \scriptsize{1.49} & 80.23 $\pm$ \scriptsize{1.13} & 82.65 $\pm$ \scriptsize{5.39} & 51.31 $\pm$ \scriptsize{1.41} \\
$D_{train-3}$ & 0.15 $\pm$ \scriptsize{0.03} & 19.03 $\pm$ \scriptsize{0.91} & 68.09 $\pm$ \scriptsize{3.64} & \textbf{85.72} $\pm$ \scriptsize{2.69} & 90.30 $\pm$ \scriptsize{1.17} & \textbf{52.66} $\pm$ \scriptsize{1.45} \\
$D_{train-4}$ & 0.17 $\pm$ \scriptsize{0.07} & 16.39 $\pm$ \scriptsize{1.55} & 63.76 $\pm$ \scriptsize{6.47} & 83.58 $\pm$ \scriptsize{4.32} & \textbf{91.54} $\pm$ \scriptsize{3.12} & 51.09 $\pm$ \scriptsize{3.00} \\

            \bottomrule
        \end{tabular}}
    \end{subtable}

    \vspace{6pt}
    
    \begin{subtable}{\linewidth}
        \centering
        \caption{Qwen-2-7B ($N_{train}=60$)}
        \resizebox{\linewidth}{!}{
        \begin{tabular}{l|ccccc|c}
            \toprule
            \textbf{Train Data} & $\mathbf{ACC_{test-0}}$ & $\mathbf{ACC_{test-1}}$ & $\mathbf{ACC_{test-2}}$ & $\mathbf{ACC_{test-3}}$ & $\mathbf{ACC_{test-4}}$ & $\mathbf{ACC_{test}}$ \\ 
            \midrule
$D_{train-0}$ & \textbf{7.85} $\pm$ \scriptsize{0.67} & 34.88 $\pm$ \scriptsize{1.29} & 57.84 $\pm$ \scriptsize{4.28} & 66.17 $\pm$ \scriptsize{6.57} & 72.70 $\pm$ \scriptsize{8.35} & 47.89 $\pm$ \scriptsize{3.99} \\
$D_{train-1}$ & 6.49 $\pm$ \scriptsize{0.61} & \textbf{39.40} $\pm$ \scriptsize{2.10} & 60.68 $\pm$ \scriptsize{4.01} & 69.50 $\pm$ \scriptsize{5.15} & 70.57 $\pm$ \scriptsize{2.88} & 49.33 $\pm$ \scriptsize{2.23} \\
$D_{train-2}$ & 0.36 $\pm$ \scriptsize{0.09} & 26.15 $\pm$ \scriptsize{1.76} & {80.16} $\pm$ \scriptsize{3.59} & 83.14 $\pm$ \scriptsize{4.35} & 85.30 $\pm$ \scriptsize{7.44} & {55.02} $\pm$ \scriptsize{3.12} \\
$D_{train-3}$ & 0.48 $\pm$ \scriptsize{0.09} & 21.58 $\pm$ \scriptsize{0.54} & \textbf{82.01} $\pm$ \scriptsize{1.89} & \textbf{92.34} $\pm$ \scriptsize{1.00} & {95.48} $\pm$ \scriptsize{4.82} & \textbf{58.38} $\pm$ \scriptsize{1.45} \\
$D_{train-4}$ & 0.21 $\pm$ \scriptsize{0.04} & 19.55 $\pm$ \scriptsize{1.02} & {80.48} $\pm$ \scriptsize{1.24} & {90.84} $\pm$ \scriptsize{1.98} & \textbf{96.00} $\pm$ \scriptsize{0.99} & {57.42} $\pm$ \scriptsize{0.92} \\

            \bottomrule
        \end{tabular}}
    \end{subtable}%
    
    \label{tbl:bucket_id}
\end{table}

\begin{table}[!t]
    \centering
\caption{Performance (out-of-domain) of LLMs trained using data with different memory levels on test sets with different memory levels. The best performance of each test set is \textbf{bolded}.}
    
    \begin{subtable}{\linewidth}
        \centering
        \caption{LLaMA-2-7B ($N_{train}=60$)}
        \resizebox{\linewidth}{!}{
        \begin{tabular}{l|ccccc|c}
            \toprule
            \textbf{Train Data} & $\mathbf{ACC_{test-ood-0}}$ & $\mathbf{ACC_{test-ood-1}}$ & $\mathbf{ACC_{test-ood-2}}$ & $\mathbf{ACC_{test-ood-3}}$ & $\mathbf{ACC_{test-ood-4}}$ & $\mathbf{ACC_{test-ood}}$ \\ 
            \midrule
$D_{train-0}$ & 0.17 $\pm$ \scriptsize{0.15} & 14.74 $\pm$ \scriptsize{4.64} & 59.60 $\pm$ \scriptsize{9.41} & 66.53 $\pm$ \scriptsize{7.08} & 64.29 $\pm$ \scriptsize{11.57} & 41.07 $\pm$ \scriptsize{6.43} \\
$D_{train-1}$ & \textbf{0.47} $\pm$ \scriptsize{0.26} & \textbf{21.13} $\pm$ \scriptsize{2.56} & \textbf{72.96} $\pm$ \scriptsize{2.13} & \textbf{78.1}3 $\pm$ \scriptsize{0.92} & \textbf{74.29} $\pm$ \scriptsize{4.30} & \textbf{49.39} $\pm$ \scriptsize{1.82} \\
$D_{train-2}$ & 0.15 $\pm$ \scriptsize{0.11} & 13.79 $\pm$ \scriptsize{0.96} & 61.48 $\pm$ \scriptsize{3.60} & 72.66 $\pm$ \scriptsize{3.23} & 71.07 $\pm$ \scriptsize{5.84} & 43.83 $\pm$ \scriptsize{2.55} \\
$D_{train-3}$ & 0.11 $\pm$ \scriptsize{0.16} & 9.85 $\pm$ \scriptsize{3.13} & 54.24 $\pm$ \scriptsize{7.35} & 70.58 $\pm$ \scriptsize{7.54} & 62.86 $\pm$ \scriptsize{9.89} & 39.53 $\pm$ \scriptsize{5.24} \\
$D_{train-4}$ & 0.07 $\pm$ \scriptsize{0.03} & 10.65 $\pm$ \scriptsize{1.90} & 59.17 $\pm$ \scriptsize{4.80} & 74.73 $\pm$ \scriptsize{3.05} & 65.36 $\pm$ \scriptsize{6.63} & 42.00 $\pm$ \scriptsize{3.22} \\

            \bottomrule
        \end{tabular}}
    \end{subtable}%

    \vspace{6pt}
    \begin{subtable}{\linewidth}
        \centering
        \caption{LLaMA-2-13B ($N_{train}=60$)}
        \resizebox{\linewidth}{!}{
        \begin{tabular}{l|ccccc|c}
            \toprule
            \textbf{Train Data} & $\mathbf{ACC_{test-ood-0}}$ & $\mathbf{ACC_{test-ood-1}}$ & $\mathbf{ACC_{test-ood-2}}$ & $\mathbf{ACC_{test-ood-3}}$ & $\mathbf{ACC_{test-ood-4}}$ & $\mathbf{ACC_{test-ood}}$ \\ 
            \midrule
$D_{train-0}$ & 0.14 $\pm$ \scriptsize{0.12} & 8.49 $\pm$ \scriptsize{3.44} & 33.28 $\pm$ \scriptsize{6.66} & 45.71 $\pm$ \scriptsize{11.23} & 66.09 $\pm$ \scriptsize{12.78} & 30.74 $\pm$ \scriptsize{5.77} \\
$D_{train-1}$ & \textbf{0.25} $\pm$ \scriptsize{0.16} & \textbf{13.81} $\pm$ \scriptsize{5.95} & \textbf{45.60} $\pm$ \scriptsize{14.56} & 58.37 $\pm$ \scriptsize{16.89} & 72.75 $\pm$ \scriptsize{18.03} & 38.16 $\pm$ \scriptsize{10.90} \\
$D_{train-2}$ & 0.19 $\pm$ \scriptsize{0.17} & 9.94 $\pm$ \scriptsize{4.32} & 44.45 $\pm$ \scriptsize{9.70} & \textbf{64.20} $\pm$ \scriptsize{9.92} & \textbf{77.97} $\pm$ \scriptsize{8.96} & \textbf{39.35} $\pm$ \scriptsize{6.38} \\
$D_{train-3}$ & 0.08 $\pm$ \scriptsize{0.05} & 7.18 $\pm$ \scriptsize{2.54} & 38.85 $\pm$ \scriptsize{7.05} & 61.13 $\pm$ \scriptsize{7.99} & 72.17 $\pm$ \scriptsize{7.20} & 35.89 $\pm$ \scriptsize{4.87} \\
$D_{train-4}$ & 0.04 $\pm$ \scriptsize{0.04} & 4.06 $\pm$ \scriptsize{2.85} & 27.78 $\pm$ \scriptsize{11.64} & 49.94 $\pm$ \scriptsize{14.81} & 52.46 $\pm$ \scriptsize{11.10} & 26.86 $\pm$ \scriptsize{7.95} \\
            \bottomrule
        \end{tabular}}
    \end{subtable}
    
    \vspace{6pt}

    \begin{subtable}{\linewidth}
        \centering
        \caption{LLaMA-3-8B ($N_{train}=60$)}
        \resizebox{\linewidth}{!}{
        \begin{tabular}{l|ccccc|c}
            \toprule
            \textbf{Train Data} & $\mathbf{ACC_{test-ood-0}}$ & $\mathbf{ACC_{test-ood-1}}$ & $\mathbf{ACC_{test-ood-2}}$ & $\mathbf{ACC_{test-ood-3}}$ & $\mathbf{ACC_{test-ood-4}}$ & $\mathbf{ACC_{test-ood}}$ \\ 
            \midrule
$D_{train-0}$ & 0.29 $\pm$ \scriptsize{0.16} & 9.54 $\pm$ \scriptsize{2.37} & 40.26 $\pm$ \scriptsize{6.03} & 54.97 $\pm$ \scriptsize{9.15} & 55.08 $\pm$ \scriptsize{9.14} & 32.03 $\pm$ \scriptsize{5.02} \\
$D_{train-1}$ & \textbf{0.50} $\pm$ \scriptsize{0.13} & \textbf{14.11} $\pm$ \scriptsize{2.12} & 53.69 $\pm$ \scriptsize{3.93} & 71.36 $\pm$ \scriptsize{2.83} & 65.69 $\pm$ \scriptsize{5.37} & 41.07 $\pm$ \scriptsize{2.08} \\
$D_{train-2}$ & 0.41 $\pm$ \scriptsize{0.10} & 13.74 $\pm$ \scriptsize{1.66} & \textbf{55.22} $\pm$ \scriptsize{3.08} & \textbf{74.24} $\pm$ \scriptsize{1.75} & 68.15 $\pm$ \scriptsize{5.29} & \textbf{42.35} $\pm$ \scriptsize{2.23} \\
$D_{train-3}$ & 0.28 $\pm$ \scriptsize{0.12} & 11.56 $\pm$ \scriptsize{1.85} & 53.49 $\pm$ \scriptsize{1.80} & 73.04 $\pm$ \scriptsize{1.30} & \textbf{68.62} $\pm$ \scriptsize{5.20} & 41.40 $\pm$ \scriptsize{1.98} \\
$D_{train-4}$ & 0.23 $\pm$ \scriptsize{0.06} & 9.95 $\pm$ \scriptsize{1.99} & 48.66 $\pm$ \scriptsize{5.98} & 67.88 $\pm$ \scriptsize{10.80} & 64.15 $\pm$ \scriptsize{9.24} & 38.17 $\pm$ \scriptsize{5.49} \\

            \bottomrule
        \end{tabular}}
    \end{subtable}

    \vspace{6pt}
    
    \begin{subtable}{\linewidth}
        \centering
        \caption{Qwen-2-7B ($N_{train}=60$)}
        \resizebox{\linewidth}{!}{
        \begin{tabular}{l|ccccc|c}
            \toprule
            \textbf{Train Data} & $\mathbf{ACC_{test-ood-0}}$ & $\mathbf{ACC_{test-ood-1}}$ & $\mathbf{ACC_{test-ood-2}}$ & $\mathbf{ACC_{test-ood-3}}$ & $\mathbf{ACC_{test-ood-4}}$ & $\mathbf{ACC_{test-ood}}$ \\ 
            \midrule
$D_{train-0}$ & 0.73 $\pm$ \scriptsize{0.09} & 19.22 $\pm$ \scriptsize{0.90} & 60.06 $\pm$ \scriptsize{1.92} & 71.81 $\pm$ \scriptsize{3.03} & 64.67 $\pm$ \scriptsize{8.69} & 43.30 $\pm$ \scriptsize{2.79} \\
$D_{train-1}$ & \textbf{0.88} $\pm$ \scriptsize{0.10} & \textbf{20.00} $\pm$ \scriptsize{0.32} & 61.91 $\pm$ \scriptsize{1.74} & 69.24 $\pm$ \scriptsize{2.71} & 68.00 $\pm$ \scriptsize{3.80} & 44.00 $\pm$ \scriptsize{0.62} \\
$D_{train-2}$ & 0.61 $\pm$ \scriptsize{0.24} & 19.89 $\pm$ \scriptsize{1.28} & 68.01 $\pm$ \scriptsize{0.51} & 79.72 $\pm$ \scriptsize{1.64} & 80.67 $\pm$ \scriptsize{2.79} & 49.78 $\pm$ \scriptsize{0.65} \\
$D_{train-3}$ & 0.37 $\pm$ \scriptsize{0.05} & 17.56 $\pm$ \scriptsize{0.32} & \textbf{70.27} $\pm$ \scriptsize{0.70} & \textbf{85.69} $\pm$ \scriptsize{1.19} & \textbf{88.00} $\pm$ \scriptsize{1.83} & \textbf{52.38} $\pm$ \scriptsize{0.70} \\
$D_{train-4}$ & 0.21 $\pm$ \scriptsize{0.06} & 13.17 $\pm$ \scriptsize{1.33} & 62.62 $\pm$ \scriptsize{4.16} & 79.17 $\pm$ \scriptsize{5.17} & 74.67 $\pm$ \scriptsize{12.16} & 45.97 $\pm$ \scriptsize{4.40} \\

            \bottomrule
        \end{tabular}}
    \end{subtable}%
    
    \label{tbl:bucket_ood}
\end{table}

\section{Further Studies}
\label{sec:further}

In this section, we further explore Q3, i.e., how do data requirements for the SFT stage vary across LLMs? On one hand, we compare the distributions of knowledge memory levels across different LLMs to examine the variations in their knowledge memorization (Section~\ref{sec:distribution}). On the other hand, we train different models using the same data to highlight the specificity of fine-tuned data for each LLM (Section~\ref{sec:compare}).

\subsection{Distribution of Knowledge Memory Levels across Different LLMs}
\label{sec:distribution}

\begin{figure}[!t]
    \centering
        \includegraphics[width=\linewidth]{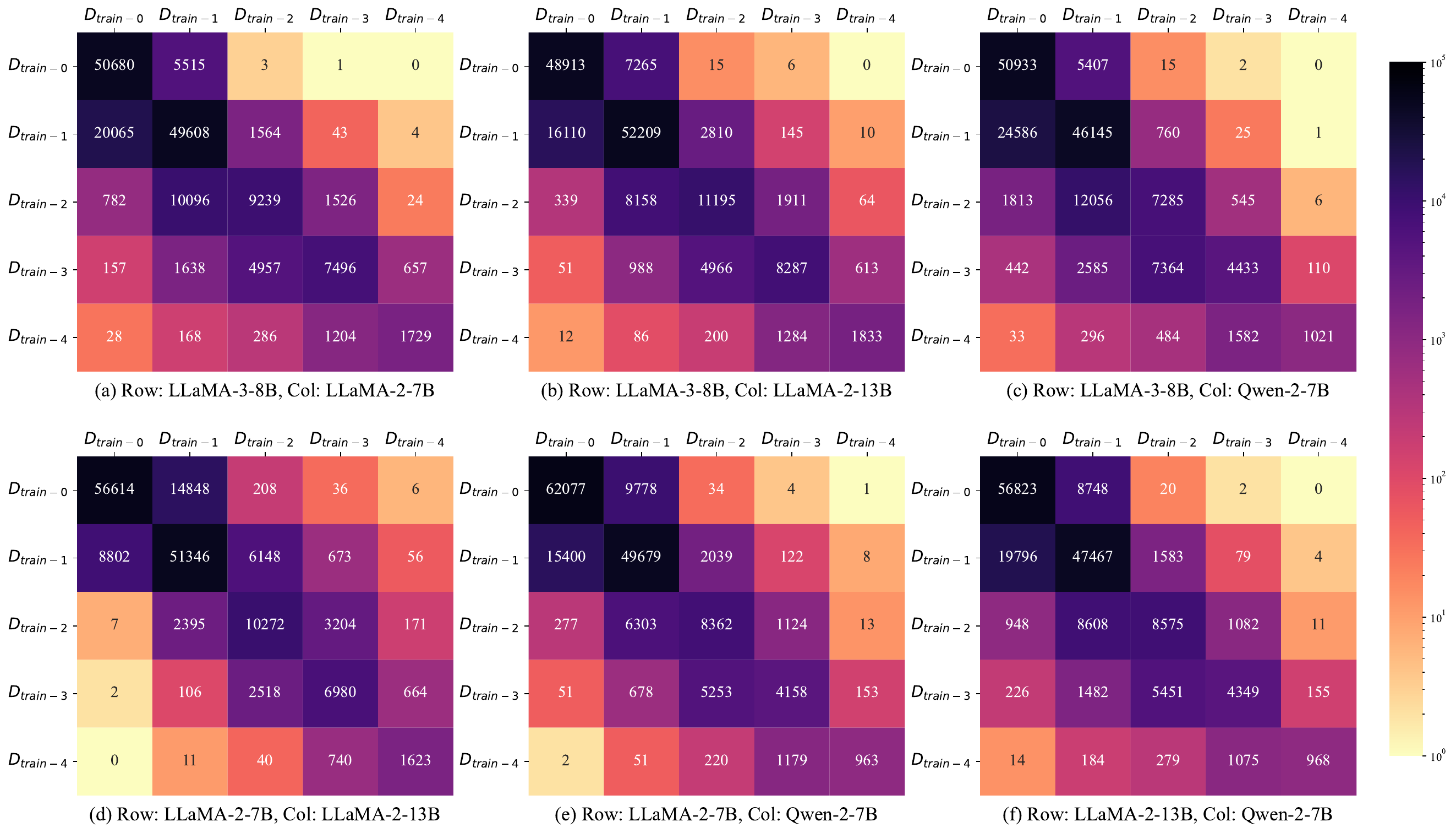}
    \caption{Heat maps showing differences in the distribution of memory levels for different LLMs on the training data $D_{train}$.}
    \label{fig:match_matrix}
\end{figure}

To thoroughly analyze the differences in knowledge memory levels among individual LLMs, we compare the memory levels of different LLMs on the training data \( D_{train} \) n a pairwise manner, with the results presented in Figure~\ref{fig:match_matrix}.

\paragraph{Finding 5}
The results in Figure~\ref{fig:match_matrix} demonstrate significant differences in knowledge distribution among various LLMs. For instance, in the heat map (c), knowledge that is difficult to memorize in Qwen-2-7B (i.e., \( D_{train-0} \)) is still partially memorized in LLaMA-3-8B, where 33 items are deeply memorized (i.e., \( D_{train-4} \)). Additionally, LLaMA-3-8B exhibits a higher level of memorization compared to other models, suggesting a broader knowledge base. Considering that the ability of LLMs to encode knowledge is linked to the amount of corresponding data in the pre-training corpus~\citep{Kandpal2022LargeLM, Part3-3}, these differences likely stem from variations in the pre-training corpora of different LLMs.

\subsection{Varying Data Requirements across LLMs}
\label{sec:compare}

\begin{table}[!t]
\centering
\caption{Performance comparison of fine-tuning different LLMs using the same 60 data samples $D_{train}^*$. ${ACC}_{test}^\dagger $ represents the best performance for each LLM trained with 60 samples.}
\resizebox{\linewidth}{!}
{
\begin{tabular}{l|ccccc|c|c}
\toprule
\textbf{Models}  & $\mathbf{ACC_{test-0}}$ & $\mathbf{ACC_{test-1}}$ & $\mathbf{ACC_{test-2}}$ & $\mathbf{ACC_{test-3}}$ & $\mathbf{ACC_{test-4}}$ & $\mathbf{ACC_{test}}$ & $\mathbf{{ACC}_{test}^\dagger}$ \\ 
\midrule
LLaMA-2-7B & 1.05 & 20.07 & 52.90 & 56.59 & 72.15 & $40.55_{\downarrow 14.30}$ & 54.85\\
LLaMA-3-8B & 0.20 & 19.91 & \textbf{67.90} & \textbf{78.91} & \textbf{89.22} & $\textbf{51.23}_{\downarrow 1.43}$ & 52.66\\
Qwen-2-7B  & \textbf{3.09} & \textbf{25.89} & 57.59 & 66.07 & 72.17 & $44.96_{\downarrow 13.42}$ & 58.38\\ 
\bottomrule

\end{tabular}
}
\label{tbl:bucket_cross}
\end{table}

Given the significant differences in knowledge memory level distribution across various LLMs, we hypothesize that the appropriate SFT data required for each LLM differs. To test this hypothesis, we train different models using the same batch of SFT data. Specifically, we select 60 data samples, which may be classified into different memory levels by different base models, denoted as \( D_{train}^* \), and use them to train LLMs from three families. The results are shown in Table~\ref{tbl:bucket_cross}.

\paragraph{Finding 6}  
The results in Table~\ref{tbl:bucket_cross} clearly illustrate the significant differences that arise when fine-tuning different LLMs using the same data. Specifically, while LLaMA-3-8B demonstrates superior performance when fine-tuned with \( D_{train}^* \), the QA abilities of the other models, particularly Qwen-2-7B, are not well-developed. Combined with the findings in Figure~\ref{fig:match_matrix}, we observe that LLMs with similar distributions of knowledge memory levels (e.g., LLaMA-2-7B and Qwen-2-7B) perform more consistently after fine-tuning with the same data. This suggests that selecting the most appropriate training data for different models should be based on their memory level distribution characteristics.

\section{Related Work}
\paragraph{Supervised Fine-Tuning}
SFT, also known as instruction tuning, is a crucial stage in the training of LLMs, primarily aimed at enhancing their performance across various downstream tasks through multi-task training~\citep{InstructGPT, analysis-chen, Zhang_IT, agent-flan}. Numerous studies~\citep{Sun_Amuro, icml/GhoshEKSAJDM24, ToolEyes, ToolSword} have examined ways to optimize SFT, highlighting that data selection during this phase significantly influences training effectiveness~\citep{DBLP:journals/nature/ShumailovSZPAG24,  acl/DongYLLXL00ZZ24, icml/XiaMGA024}. For example, both \cite{nips/ZhouLX0SMMEYYZG23} and \cite{Liu_What} demonstrated that LLMs can be better tuned for general tasks by using fewer but higher-quality instruction-tuned data. In this work, we investigate the mechanisms underlying SFT in the context of the QA task and explore strategies for selecting more effective SFT data.

\paragraph{Question-Answering}
The QA task requires LLMs to effectively utilize the knowledge encoded during pre-training to answer users' questions~\citep{QA-1, QA-2, QA-3}. However, LLMs can sometimes provide incorrect answers to factual questions~\citep{Huang_hall}. Some researchers~\citep{Kandpal2022LargeLM, emnlp/KangC23} attribute this to the underutilization of pre-training data, leading to knowledge shortcuts and recall failures, while others~\citep{Schulman_youtube} suggest that capability misalignment introduced during post-training are to blame. Regardless, enhancing the QA capabilities of LLMs remains a key concern for researchers. \cite{sft-renmin} indicated that SFT for the QA task should maintain consistency in model knowledge before and after fine-training. Meanwhile, \cite{sft-google} focused on hallucinations caused by new knowledge training and found that poorly memorized knowledge in pre-trained LLMs significantly increases these hallucinations. Our study aims to shed light on the role of SFT in the QA task and inform the development of more effective SFT strategies. We investigate the impact of data volume during the SFT stage, introduce a more precise multi-template complementation mechanism to evaluate knowledge memorization levels, and perform finer data segmentation to obtain more accurate and detailed results.

\section{Conclusion and Future Work}

In this paper, we present a comprehensive empirical analysis of fine-tuning LLMs for the QA task. We propose a memory discrimination approach based on a multi-template complementation mechanism to thoroughly explore the data requirements for SFT, the effects of using data with varying memory levels, and the differences in data needs across LLMs. We hope our findings will provide valuable insights for designing more effective SFT strategies.

In the future, we aim to build on these results by conducting deeper analyses of the underlying mechanisms in LLMs. Our objective is to investigate performance variations in LLMs caused by SFT under different conditions and the resulting model changes, with the goal of clarifying key characteristics of using LLMs for the QA task.

\bibliographystyle{abbrvnat}
\bibliography{main}

\clearpage
\appendix

\section{Details of Data}
\label{sec:detail_data}

We categorize 12 location-related topics as in-domain data and another 12 unrelated topics as out-of-domain data, designing 21 mapping templates for each topic. The corresponding data details of in-domain data are listed from Table~\ref{tab:P17} to Table~\ref{tab:P740}, while the corresponding data details of out-of-domain data are listed from Table~\ref{tab:P112} to Table~\ref{tab:P800}.

\section{Prompt Template for Each LLM}
\label{sec:prompt}

To achieve optimal performance from each LLM, we fine-tune the models using their officially recommended templates, as detailed in Table~\ref{tab:prompt}.

\begin{table}[H]
\centering
\caption{Information and mapping templates for topic P17 (in-domain).}
\begin{tabular}{p{0.95\linewidth}}
\toprule
\textbf{Topic:} P17 \\ \textbf{Question Template: } Which country is \{subject\} located in? \\ \textbf{Answer Template: } \{object\} \\ \midrule
\textbf{Mapping Templates:} \\ 
\{subject\} is located in \{object\}\\ 
The location of \{subject\} is in \{object\}\\ 
\{subject\} finds its place within the borders of \{object\}\\ 
The \{subject\} is situated in the country, \{object\}\\ 
If you're seeking the \{subject\}, look no further than the nation of \{object\}\\ 
The land encompassing the \{subject\} is known as \{object\}\\ 
\{subject\} can be found in \{object\}\\ 
\{subject\} has its roots in \{object\}\\ 
The place \{subject\} calls home is \{object\}\\ 
\{subject\} is situated in \{object\}\\ 
The geographical location of \{subject\} is in \{object\}\\ 
\{subject\} can be discovered in the nation of \{object\}\\ 
The country where \{subject\} is found is \{object\}\\ 
\{subject\}'s location is in \{object\}\\ 
\{subject\} resides in \{object\}\\ 
The country of \{subject\} is \{object\}\\ 
\{subject\} belongs to \{object\}\\ 
\{subject\} exists in \{object\}\\ 
You can find \{subject\} in \{object\}\\ 
\{subject\} is a part of \{object\}\\ 
\{subject\} lies within the borders of \{object\} \\ \bottomrule
\end{tabular}
\label{tab:P17}
\end{table}

\begin{table}[H]
\centering
\caption{Information and mapping templates for topic P19 (in-domain).}
\begin{tabular}{p{0.95\linewidth}}
\toprule
\textbf{Topic:} P19 \\ \textbf{Question Template: } Where was \{subject\} born? \\ \textbf{Answer Template: } \{object\} \\ \midrule
\textbf{Mapping Templates:} \\ 
\{subject\} was born in \{object\}\\ 
The birthplace of \{subject\} was \{object\}\\ 
It is known that \{subject\} came into the world in \{object\}\\ 
\{subject\} entered the world in \{object\}\\ 
\{subject\} was born, and that location is \{object\}\\ 
\{subject\}'s life began in \{object\}\\ 
The location of \{subject\}'s birth is \{object\}\\ 
\{subject\}'s birth occurred in \{object\}\\ 
The place where \{subject\} was born is \{object\}\\ 
\{subject\} hailed from \{object\}\\ 
The answer to where \{subject\} was born lies in \{object\}\\ 
\{subject\} originated from \{object\}\\ 
\{subject\} came into this world in \{object\}\\ 
\{subject\} entered life in \{object\}\\ 
\{subject\} first drew breath in \{object\}\\ 
The origin of \{subject\} is in \{object\}\\ 
\{subject\} hails from \{object\}\\ 
The place of birth for \{subject\} is \{object\}\\ 
\{subject\}'s birth took place in \{object\}\\ 
When it comes to birth, \{subject\} was born in \{object\}\\ 
If you were to ask where \{subject\} was born, it would be \{object\} \\ \bottomrule
\end{tabular}
\label{tab:P19}
\end{table}

\begin{table}[H]
\centering
\caption{Information and mapping templates for topic P20 (in-domain).}
\begin{tabular}{p{0.95\linewidth}}
\toprule
\textbf{Topic:} P20 \\ \textbf{Question Template: } Where did \{subject\} die? \\ \textbf{Answer Template: } \{object\} \\ \midrule
\textbf{Mapping Templates:} \\ 
\{subject\} met their end at \{object\}\\ 
\{subject\} breathed their last at \{object\}\\ 
\{subject\}'s life came to a close at \{object\}\\ 
The place of \{subject\}'s death is \{object\}\\ 
The location of \{subject\}'s demise is \{object\}\\ 
The site of \{subject\}'s final rest is \{object\}\\ 
The place where \{subject\} passed away is \{object\}\\ 
\{subject\}'s mortal remains are in \{object\}\\ 
\{subject\} succumbed to death in \{object\}\\ 
The destination of \{subject\}'s last days was \{object\}\\ 
The story of \{subject\}'s life concluded in \{object\}\\ 
\{subject\} bid farewell to the world from within the confines of \{object\}\\ 
The final resting place of \{subject\} is \{object\}\\ 
\{subject\} took his final breath in \{object\}\\ 
\{subject\}'s life journey ended in \{object\}\\ 
\{subject\} died in \{object\}\\ 
The place where \{subject\} died is \{object\}\\ 
\{subject\}'s death occurred in \{object\}\\ 
\{subject\} took their last breath in \{object\}\\ 
When it comes to death, \{subject\} died in \{object\}\\ 
Looking at the end of \{subject\}'s life, they died in \{object\} \\ \bottomrule
\end{tabular}
\label{tab:P20}
\end{table}

\begin{table}[H]
\centering
\caption{Information and mapping templates for topic P36 (in-domain).}
\begin{tabular}{p{0.95\linewidth}}
\toprule
\textbf{Topic:} P36 \\ \textbf{Question Template: } What is the capital of \{subject\}? \\ \textbf{Answer Template: } \{object\} \\ \midrule
\textbf{Mapping Templates:} \\ 
The capital of \{subject\} is \{object\}\\ 
When considering the capital of \{subject\}, it is \{object\}\\ 
In \{subject\}, the city designated as the capital is \{object\}\\ 
\{subject\}'s capital city is \{object\}\\ 
The capital city of \{subject\} is located in \{object\}\\ 
\{subject\} is governed from \{object\}\\ 
The seat of government in \{subject\} is \{object\}\\ 
\{subject\}'s governmental hub is \{object\}\\ 
The administrative center of \{subject\} is \{object\}\\ 
The political heart of \{subject\} beats in \{object\}\\ 
One can find \{subject\}'s seat of power in the city of \{object\}\\ 
One would find \{subject\}'s governing institutions nestled within the boundaries of \{object\}\\ 
\{subject\}'s capital is \{object\}\\ 
The capital of the region \{subject\} is \{object\}\\ 
\{subject\}'s capital designation goes to \{object\}\\ 
The main city of \{subject\} is \{object\}\\ 
\{subject\} has its capital in \{object\}\\ 
The chief city of \{subject\} is \{object\}\\ 
Looking at \{subject\}, its capital is \{object\}\\ 
In terms of capital cities, \{subject\} has \{object\}\\ 
As the capital of \{subject\}, you'll find \{object\} \\ \bottomrule
\end{tabular}
\label{tab:P36}
\end{table}

\begin{table}[H]
\centering
\caption{Information and mapping templates for topic P69 (in-domain).}
\begin{tabular}{p{0.95\linewidth}}
\toprule
\textbf{Topic:} P69 \\ \textbf{Question Template: } Where was \{subject\} educated? \\ \textbf{Answer Template: } \{object\} \\ \midrule
\textbf{Mapping Templates:} \\ 
\{subject\} received education at \{object\}\\ 
\{subject\} completed the studies at \{object\}\\ 
\{subject\} was schooled at \{object\}\\ 
\{subject\} was educated in \{object\}\\ 
\{subject\} graduated from \{object\}\\ 
\{subject\} spent the formative years at \{object\}\\ 
\{subject\}'s alma mater is \{object\}\\ 
\{subject\} pursued the studies at \{object\}\\ 
\{subject\} gained the knowledge at \{object\}\\ 
The academic journey of \{subject\} took place in \{object\}\\ 
The institution where \{subject\} studied is \{object\}\\ 
Education for \{subject\} was pursued within the walls of \{object\}\\ 
The educational institution attended by \{subject\} is \{object\}\\ 
\{subject\} is an alumnus/alumna of \{object\}\\ 
The academic background of \{subject\} includes \{object\}\\ 
The place where \{subject\} was educated is \{object\}\\ 
\{subject\} attended school in \{object\}\\ 
The education of \{subject\} took place in \{object\}\\ 
The place of \{subject\}'s education is \{object\}\\ 
\{subject\} received their education from \{object\}\\ 
In terms of education, \{subject\} was schooled in \{object\} \\ \bottomrule
\end{tabular}
\label{tab:P69}
\end{table}

\begin{table}[H]
\centering
\caption{Information and mapping templates for topic P131 (in-domain).}
\begin{tabular}{p{0.95\linewidth}}
\toprule
\textbf{Topic:} P131 \\ \textbf{Question Template: } Where is \{subject\} located? \\ \textbf{Answer Template: } \{object\} \\ \midrule
\textbf{Mapping Templates:} \\ 
The location of \{subject\} is where you'll find \{object\}\\ 
If you look where \{subject\} is, you'll see \{object\}\\ 
Where \{subject\} resides, there also is \{object\}\\ 
\{subject\} is located at \{object\}\\ 
\{subject\} can be found in \{object\}\\ 
\{subject\} is positioned at \{object\}\\ 
\{subject\} is stationed at \{object\}\\ 
\{subject\} is based at \{object\}\\ 
\{subject\} is headquartered at \{object\}\\ 
The current location of \{subject\} is \{object\}\\ 
One would locate \{subject\} in the vicinity of \{object\}\\ 
Currently, \{subject\} resides or occupies \{object\}\\ 
\{subject\} is in \{object\}\\ 
The geographical position of \{subject\} is \{object\}\\ 
\{subject\} is placed in \{object\}\\ 
You can find \{subject\} in \{object\}\\ 
\{subject\} exists in \{object\}\\ 
\{subject\} lies in \{object\}\\ 
The location of \{subject\} is \{object\}\\ 
\{subject\} is situated in \{object\}\\ 
\{subject\} resides in \{object\} \\ \bottomrule
\end{tabular}
\label{tab:P131}
\end{table}

\begin{table}[H]
\centering
\caption{Information and mapping templates for topic P159 (in-domain).}
\begin{tabular}{p{0.95\linewidth}}
\toprule
\textbf{Topic:} P159 \\ \textbf{Question Template: } Where is the headquarter of \{subject\}? \\ \textbf{Answer Template: } \{object\} \\ \midrule
\textbf{Mapping Templates:} \\ 
The headquarter of \{subject\} is located in \{object\}\\ 
\{subject\} has its headquarter in \{object\}\\ 
You can find the headquarter of \{subject\} in \{object\}\\ 
\{subject\}'s central office is situated in \{object\}\\ 
The main hub of \{subject\} is \{object\}\\ 
\{subject\} is headquartered in \{object\}\\ 
The location of \{subject\}'s headquarter is \{object\}\\ 
\{subject\}'s headquarter can be found at \{object\}\\ 
The address of \{subject\}'s headquarter is \{object\}\\ 
\{subject\}'s headquarters are located at \{object\}\\ 
The central hub of operations for \{subject\} can be found in \{object\}\\ 
The administrative heart of \{subject\} resides at \{object\}\\ 
\{subject\}'s head office is located in \{object\}\\ 
\{subject\} has its main base in \{object\}\\ 
\{subject\}'s headquarters can be found in \{object\}\\ 
The headquarters of \{subject\} is located in \{object\}\\ 
\{subject\}'s headquarters is in \{object\}\\ 
The main office of \{subject\} is in \{object\}\\ 
\{subject\}'s headquarter is located in \{object\}\\ 
The headquarter of \{subject\} is situated in \{object\}\\ 
When it comes to headquarters, \{subject\}'s is in \{object\} \\ \bottomrule
\end{tabular}
\label{tab:P159}
\end{table}

\begin{table}[H]
\centering
\caption{Information and mapping templates for topic P176 (in-domain).}
\begin{tabular}{p{0.95\linewidth}}
\toprule
\textbf{Topic:} P176 \\ \textbf{Question Template: } Which company is \{subject\} produced by? \\ \textbf{Answer Template: } \{object\} \\ \midrule
\textbf{Mapping Templates:} \\ 
\{subject\} is produced by \{object\}\\ 
The producer of \{subject\} is \{object\}\\ 
The production company behind \{subject\} is \{object\}\\ 
\{subject\} is created by \{object\}\\ 
\{subject\} is assembled by \{object\}\\ 
\{subject\} comes from \{object\}\\ 
\{subject\} is manufactured by \{object\}\\ 
The company responsible for \{subject\} is \{object\}\\ 
\{subject\} is a product of \{object\}\\ 
The production of \{subject\} falls under the umbrella of \{object\}\\ 
\{subject\} comes from the production house of \{object\}\\ 
The production of \{subject\} is handled by none other than \{object\}\\ 
The company behind the production of \{subject\} is \{object\}\\ 
The company that crafts \{subject\} is \{object\}\\ 
Every unit of \{subject\} bears the production mark of \{object\}\\ 
\{subject\} comes from the company \{object\}\\ 
The production of \{subject\} is handled by \{object\}\\ 
The company responsible for producing \{subject\} is \{object\}\\ 
The company that produces \{subject\} is \{object\}\\ 
When it comes to production, \{subject\} is produced by \{object\}\\ 
Looking at the production of \{subject\}, it is done by \{object\} \\ \bottomrule
\end{tabular}
\label{tab:P176}
\end{table}

\begin{table}[H]
\centering
\caption{Information and mapping templates for topic P276 (in-domain).}
\begin{tabular}{p{0.95\linewidth}}
\toprule
\textbf{Topic:} P276 \\ \textbf{Question Template: } Where is \{subject\} located? \\ \textbf{Answer Template: } \{object\} \\ \midrule
\textbf{Mapping Templates:} \\ 
\{subject\} can be found at \{object\}\\ 
The location of \{subject\} is \{object\}\\ 
\{subject\} is situated at \{object\}\\ 
\{subject\} has its base in \{object\}\\ 
\{subject\} is headquartered in \{object\}\\ 
\{subject\} operates out of \{object\}\\ 
The place where \{subject\} is located is \{object\}\\ 
\{subject\} is positioned at \{object\}\\ 
The site of \{subject\} is \{object\}\\ 
\{subject\} can be found in the location \{object\}\\ 
The whereabouts of \{subject\} are at \{object\}\\ 
\{subject\} is situated in the place called \{object\}\\ 
\{subject\} is established in \{object\}\\ 
The coordinates of \{subject\} point to \{object\}\\ 
The address of \{subject\} leads to \{object\}\\ 
\{subject\} is located in \{object\}\\ 
\{subject\} resides in \{object\}\\ 
You can find \{subject\} in \{object\}\\ 
When it comes to location, \{subject\} is in \{object\}\\ 
Looking at where \{subject\} is, it is in \{object\}\\ 
In terms of location, \{subject\} is situated in \{object\} \\ \bottomrule
\end{tabular}
\label{tab:P276}
\end{table}

\begin{table}[H]
\centering
\caption{Information and mapping templates for topic P413 (in-domain).}
\begin{tabular}{p{0.95\linewidth}}
\toprule
\textbf{Topic:} P413 \\ \textbf{Question Template: } What position does \{subject\} play? \\ \textbf{Answer Template: } \{object\} \\ \midrule
\textbf{Mapping Templates:} \\ 
\{subject\} plays \{object\}\\ 
The position of \{subject\} is \{object\}\\ 
In the team, \{subject\} holds the position of \{object\}\\ 
\{subject\} plays the position of \{object\}\\ 
\{subject\}'s role is \{object\}\\ 
\{subject\} is a \{object\}\\ 
The position played by \{subject\} is \{object\}\\ 
\{subject\} holds the position of \{object\}\\ 
\{subject\} is a player in the position of \{object\}\\ 
In the game, \{subject\} assumes the role of \{object\}\\ 
\{subject\} is known for the position as \{object\}\\ 
When playing, \{subject\} takes up the position of \{object\}\\ 
The role \{subject\} takes on is \{object\}\\ 
\{subject\} is assigned to the position \{object\}\\ 
The position that \{subject\} occupies is \{object\}\\ 
\{subject\} occupies the position of \{object\}\\ 
\{subject\} fulfills the role of \{object\}\\ 
\{subject\} is positioned as a \{object\}\\ 
The position that \{subject\} plays is \{object\}\\ 
\{subject\}'s position is \{object\}\\ 
If you were to ask what position \{subject\} plays, it's \{object\} \\ \bottomrule
\end{tabular}
\label{tab:P413}
\end{table}

\begin{table}[H]
\centering
\caption{Information and mapping templates for topic P495 (in-domain).}
\begin{tabular}{p{0.95\linewidth}}
\toprule
\textbf{Topic:} P495 \\ \textbf{Question Template: } Which country was \{subject\} created in? \\ \textbf{Answer Template: } \{object\} \\ \midrule
\textbf{Mapping Templates:} \\ 
\{subject\} was created in \{object\}\\ 
The creation of \{subject\} took place in \{object\}\\ 
The origin of \{subject\} is traced back to \{object\}\\ 
\{subject\} was born in \{object\}\\ 
\{subject\} originated from \{object\}\\ 
\{subject\} was founded in \{object\}\\ 
\{subject\} was created in the country of \{object\}\\ 
The country of origin for \{subject\} is \{object\}\\ 
\{subject\} originated in the country of \{object\}\\ 
The birthplace of \{subject\} is none other than \{object\}\\ 
\{subject\}'s formation occurred in the borders of \{object\}\\ 
Historically, \{subject\} emerged in the country known as \{object\}\\ 
\{subject\} was conceptualized in \{object\}\\ 
The country credit for the creation of \{subject\} goes to \{object\}\\ 
The country that witnessed the creation of \{subject\} is \{object\}\\ 
The country where \{subject\} was created is \{object\}\\ 
\{subject\} was made in \{object\}\\ 
\{subject\} came into being in \{object\}\\ 
If you were to ask where \{subject\} was created, it would be \{object\}\\ 
Looking at the origin of \{subject\}, it was created in \{object\}\\ 
In terms of country of origin, \{subject\} was created in \{object\} \\ \bottomrule
\end{tabular}
\label{tab:P495}
\end{table}

\begin{table}[H]
\centering
\caption{Information and mapping templates for topic P740 (in-domain).}
\begin{tabular}{p{0.95\linewidth}}
\toprule
\textbf{Topic:} P740 \\ \textbf{Question Template: } Where was \{subject\} founded? \\ \textbf{Answer Template: } \{object\} \\ \midrule
\textbf{Mapping Templates:} \\ 
The founding of \{subject\} took place in \{object\}\\ 
\{subject\} was originally established in \{object\}\\ 
\{subject\}'s origin is traced back to \{object\}\\ 
\{subject\} was founded in \{object\}\\ 
\{subject\} originated in \{object\}\\ 
\{subject\} has its roots in \{object\}\\ 
The founding location of \{subject\} is \{object\}\\ 
\{subject\} has its origins in \{object\}\\ 
The birthplace of \{subject\} is \{object\}\\ 
One can trace \{subject\}'s beginnings to \{object\}\\ 
\{subject\} came into existence in \{object\}\\ 
The roots of \{subject\} dig deep into the soil of \{object\}\\ 
\{subject\} traces its beginnings back to \{object\}\\ 
The inception of \{subject\} took place in \{object\}\\ 
\{subject\} was brought into existence in \{object\}\\ 
The founding place of \{subject\} is \{object\}\\ 
The origin of \{subject\} is in \{object\}\\ 
The establishment of \{subject\} occurred in \{object\}\\ 
If you were to ask where \{subject\} was founded, it would be \{object\}\\ 
Looking at the origin of \{subject\}, it was founded in \{object\}\\ 
In terms of its founding location, \{subject\} was established in \{object\} \\ \bottomrule
\end{tabular}
\label{tab:P740}
\end{table}

\begin{table}[H]
\centering
\caption{Information and mapping templates for topic P112 (out-of-domain).}
\begin{tabular}{p{0.95\linewidth}}
\toprule
\textbf{Topic:} P112 \\ \textbf{Question Template: } Who founded \{subject\}? \\ \textbf{Answer Template: } \{object\} \\ \midrule
\textbf{Mapping Templates:} \\ 
The founder of \{subject\} is \{object\}\\ 
\{subject\} was founded by \{object\}\\ 
The establishment of \{subject\} was initiated by \{object\}\\ 
\{subject\} owes its existence to \{object\}\\ 
\{subject\} was brought into being by \{object\}\\ 
\{subject\} is a brainchild of \{object\}\\ 
\{subject\} was established by \{object\}\\ 
\{subject\} has its roots in \{object\}\\ 
The person who founded \{subject\} is \{object\}\\ 
The visionary behind \{subject\}'s establishment is \{object\}\\ 
The inception of \{subject\} can be traced back to \{object\}\\ 
The idea and realization of \{subject\} were the brainchild of \{object\}\\ 
\{subject\} was brought into existence by \{object\}\\ 
\{subject\}'s founder is known to be \{object\}\\ 
\{subject\} owes its inception to \{object\}\\ 
\{subject\} was created by \{object\}\\ 
The creation of \{subject\} is attributed to \{object\}\\ 
\{subject\} was started by \{object\}\\ 
\{subject\} originated with \{object\}\\ 
\{subject\}'s origins lie with \{object\}\\ 
\{subject\} can trace its roots back to \{object\} \\ \bottomrule
\end{tabular}
\label{tab:P112}
\end{table}

\begin{table}[H]
\centering
\caption{Information and mapping templates for topic P127 (out-of-domain).}
\begin{tabular}{p{0.95\linewidth}}
\toprule
\textbf{Topic:} P127 \\ \textbf{Question Template: } Who owns \{subject\}? \\ \textbf{Answer Template: } \{object\} \\ \midrule
\textbf{Mapping Templates:} \\ 
The owner of \{subject\} is \{object\}\\ 
\{subject\} is owned by \{object\}\\ 
Ownership of \{subject\} belongs to \{object\}\\ 
\{subject\} belongs to \{object\}\\ 
\{subject\} is in the possession of \{object\}\\ 
\{subject\} is a property of \{object\}\\ 
\{subject\} is possessed by \{object\}\\ 
\{subject\} is under the ownership of \{object\}\\ 
\{subject\} is held by \{object\}\\ 
The proprietor of \{subject\} is none other than \{object\}\\ 
Responsibility for \{subject\} falls under the jurisdiction of \{object\}\\ 
The property known as \{subject\} is under the stewardship of \{object\}\\ 
The rights to \{subject\} are held by \{object\}\\ 
The individual who owns \{subject\} is \{object\}\\ 
The rightful owner of \{subject\} is identified as \{object\}\\ 
Ownership of \{subject\} is held by \{object\}\\ 
The possession of \{subject\} is with \{object\}\\ 
The entity owning \{subject\} is \{object\}\\ 
\{subject\}'s owner is \{object\}\\ 
\{subject\} is in the hands of \{object\}\\ 
If you're looking for the owner of \{subject\}, it's \{object\} \\ \bottomrule
\end{tabular}
\label{tab:P127}
\end{table}

\begin{table}[H]
\centering
\caption{Information and mapping templates for topic P170 (out-of-domain).}
\begin{tabular}{p{0.95\linewidth}}
\toprule
\textbf{Topic:} P170 \\ \textbf{Question Template: } Who was \{subject\} created by? \\ \textbf{Answer Template: } \{object\} \\ \midrule
\textbf{Mapping Templates:} \\ 
\{subject\} was created by \{object\}\\ 
The creator of \{subject\} was \{object\}\\ 
The person who created \{subject\} is known as \{object\}\\ 
\{subject\} was founded by \{object\}\\ 
\{subject\} owes its creation to \{object\}\\ 
\{subject\} was developed by \{object\}\\ 
\{subject\}'s creator is \{object\}\\ 
\{subject\} was the creation of \{object\}\\ 
The person behind \{subject\} is \{object\}\\ 
\{subject\} was brought into existence by \{object\}\\ 
The originator of \{subject\} is \{object\}\\ 
The creative force behind \{subject\} is attributed to \{object\}\\ 
\{subject\} came into existence thanks to \{object\}\\ 
\{subject\} was brought to life by \{object\}\\ 
\{subject\} was conceptualized by \{object\}\\ 
The creation of \{subject\} is attributed to \{object\}\\ 
The entity responsible for creating \{subject\} is \{object\}\\ 
\{subject\} was made by \{object\}\\ 
\{subject\}'s creation is attributed to \{object\}\\ 
When it comes to creation, \{subject\} was created by \{object\}\\ 
Looking at the creation of \{subject\}, it was done by \{object\} \\ \bottomrule
\end{tabular}
\label{tab:P170}
\end{table}

\begin{table}[H]
\centering
\caption{Information and mapping templates for topic P175 (out-of-domain).}
\begin{tabular}{p{0.95\linewidth}}
\toprule
\textbf{Topic:} P175 \\ \textbf{Question Template: } Who performed \{subject\}? \\ \textbf{Answer Template: } \{object\} \\ \midrule
\textbf{Mapping Templates:} \\ 
The performer of \{subject\} was \{object\}\\ 
\{subject\} was performed by \{object\}\\ 
The one responsible for performing \{subject\} was \{object\}\\ 
\{subject\} was brought to life by \{object\}\\ 
\{subject\} was presented by \{object\}\\ 
\{subject\} was executed by \{object\}\\ 
The artist behind \{subject\} is \{object\}\\ 
The talent behind \{subject\} is \{object\}\\ 
The one who performed \{subject\} was \{object\}\\ 
The one who executed \{subject\} skillfully was \{object\}\\ 
The artist responsible for \{subject\}'s interpretation was \{object\}\\ 
The responsibility of performing \{subject\} fell upon \{object\}\\ 
\{subject\} was enacted by \{object\}\\ 
The act of \{subject\} was performed by \{object\}\\ 
\{subject\} was executed on stage by \{object\}\\ 
The execution of \{subject\} was done by \{object\}\\ 
\{subject\} was carried out by \{object\}\\ 
The realization of \{subject\} was by \{object\}\\ 
\{subject\} had its performance by \{object\}\\ 
The performance of \{subject\} was done by \{object\}\\ 
Looking at the performance of \{subject\}, it was done by \{object\} \\ \bottomrule
\end{tabular}
\label{tab:P175}
\end{table}

\begin{table}[H]
\centering
\caption{Information and mapping templates for topic P26 (out-of-domain).}
\begin{tabular}{p{0.95\linewidth}}
\toprule
\textbf{Topic:} P26 \\ \textbf{Question Template: } Who is \{subject\} married to? \\ \textbf{Answer Template: } \{object\} \\ \midrule
\textbf{Mapping Templates:} \\ 
\{subject\}'s spouse is \{object\}\\ 
\{subject\} has been married to \{object\}\\ 
\{subject\} is in a marital union with \{object\}\\ 
The person \{subject\} is married to is \{object\}\\ 
\{subject\}'s partner in marriage is \{object\}\\ 
\{subject\}'s better half is \{object\}\\ 
\{subject\} is wed to \{object\}\\ 
\{subject\} exchanged vows with \{object\}\\ 
\{subject\} shares a life with \{object\}\\ 
\{subject\} shares a marital bond with \{object\}\\ 
Their love story culminated in a wedding, uniting \{subject\} and \{object\}\\ 
The answer to \{subject\}'s nuptials lies in the presence of \{object\}\\ 
\{subject\} is married to \{object\}\\ 
\{subject\} has tied the knot with \{object\}\\ 
\{subject\} shares a matrimonial life with \{object\}\\ 
The spouse of \{subject\} is \{object\}\\ 
\{subject\} is wedded to \{object\}\\ 
In marriage, \{subject\} is united with \{object\}\\ 
The one \{subject\} is married to is \{object\}\\ 
\{subject\}'s husband/wife is \{object\}\\ 
When it comes to marriage, \{subject\} is married to \{object\} \\ \bottomrule
\end{tabular}
\label{tab:P26}
\end{table}

\begin{table}[H]
\centering
\caption{Information and mapping templates for topic P40 (out-of-domain).}
\begin{tabular}{p{0.95\linewidth}}
\toprule
\textbf{Topic:} P40 \\ \textbf{Question Template: } Who is \{subject\}'s child? \\ \textbf{Answer Template: } \{object\} \\ \midrule
\textbf{Mapping Templates:} \\ 
The child of \{subject\} is known to be \{object\}\\ 
Belonging to \{subject\} as a child is \{object\}\\ 
As a child to \{subject\}, there is \{object\}\\ 
\{subject\}'s child is \{object\}\\ 
\{subject\} is the parent of \{object\}\\ 
\{subject\}'s offspring is \{object\}\\ 
\{subject\}'s youngster is \{object\}\\ 
\{subject\}'s family includes \{object\}\\ 
\{subject\}'s lineage includes \{object\}\\ 
\{subject\} has a child named \{object\}\\ 
The offspring of \{subject\} is identified as \{object\}\\ 
The child of \{subject\} is recognized as \{object\}\\ 
The offspring of \{subject\} includes \{object\}\\ 
\{subject\} is the biological parent of \{object\}\\ 
\{subject\} is the father/mother to \{object\}\\ 
The child of \{subject\} is \{object\}\\ 
The progeny of \{subject\} is \{object\}\\ 
The next generation of \{subject\} includes \{object\}\\ 
If you were to ask who \{subject\}'s child is, it's \{object\}\\ 
Looking at \{subject\}'s offspring, it's \{object\}\\ 
In terms of children, \{subject\} has \{object\} \\ \bottomrule
\end{tabular}
\label{tab:P40}
\end{table}

\begin{table}[H]
\centering
\caption{Information and mapping templates for topic P50 (out-of-domain).}
\begin{tabular}{p{0.95\linewidth}}
\toprule
\textbf{Topic:} P50 \\ \textbf{Question Template: } Who is the author of \{subject\}? \\ \textbf{Answer Template: } \{object\} \\ \midrule
\textbf{Mapping Templates:} \\ 
\{subject\} was authored by \{object\}\\ 
The writer of \{subject\} is \{object\}\\ 
The person who authored \{subject\} is \{object\}\\ 
The author of \{subject\} is \{object\}\\ 
\{subject\} was written by \{object\}\\ 
\{subject\} is a work by \{object\}\\ 
The creator of \{subject\} is \{object\}\\ 
The person responsible for \{subject\} is \{object\}\\ 
\{subject\} owes its existence to \{object\}\\ 
The creative mind behind \{subject\} is none other than \{object\}\\ 
\{subject\} was penned by the talented writer, \{object\}\\ 
The work known as \{subject\} was brought to life by the author, \{object\}\\ 
\{subject\} is a work authored by \{object\}\\ 
The penname associated with \{subject\} is \{object\}\\ 
The words of \{subject\} were put together by \{object\}\\ 
The person who wrote \{subject\} is \{object\}\\ 
\{subject\} was created by \{object\}\\ 
\{subject\} was drafted by \{object\}\\ 
If you were to ask who authored \{subject\}, it was \{object\}\\ 
Looking at the authorship of \{subject\}, it was written by \{object\}\\ 
\{subject\} is a creation of \{object\} \\ \bottomrule
\end{tabular}
\label{tab:P50}
\end{table}

\begin{table}[H]
\centering
\caption{Information and mapping templates for topic P136 (out-of-domain).}
\begin{tabular}{p{0.95\linewidth}}
\toprule
\textbf{Topic:} P136 \\ \textbf{Question Template: } What type of music does \{subject\} play? \\ \textbf{Answer Template: } \{object\} \\ \midrule
\textbf{Mapping Templates:} \\ 
The music played by \{subject\} is \{object\}\\ 
When \{subject\} plays music, it is \{object\}\\ 
The musical style of \{subject\} can be categorized as \{object\}\\ 
\{subject\}'s sound is characterized as \{object\}\\ 
\{subject\}'s musical talent lies in \{object\}\\ 
\{subject\} has a knack for \{object\}\\ 
\{subject\}'s genre of music is \{object\}\\ 
\{subject\} is known for playing \{object\}\\ 
\{subject\}'s music style is \{object\}\\ 
The genre that \{subject\} excels in is \{object\}\\ 
When it comes to music, \{subject\} is known for their proficiency in \{object\}\\ 
The tunes produced by \{subject\} belong to the category of \{object\}\\ 
\{subject\}'s music falls under the category of \{object\}\\ 
\{subject\} has a musical style that is categorized as \{object\}\\ 
The music played by \{subject\} can be described as \{object\}\\ 
The type of music \{subject\} plays is \{object\}\\ 
The genre of music \{subject\} plays is \{object\}\\ 
The style of music \{subject\} plays is \{object\}\\ 
\{subject\} plays the music type of \{object\}\\ 
Musically, \{subject\} is known to play \{object\}\\ 
In terms of musical style, \{subject\} plays \{object\} \\ \bottomrule
\end{tabular}
\label{tab:P136}
\end{table}

\begin{table}[H]
\centering
\caption{Information and mapping templates for topic P106 (out-of-domain).}
\begin{tabular}{p{0.95\linewidth}}
\toprule
\textbf{Topic:} P106 \\ \textbf{Question Template: } What kind of work does \{subject\} do? \\ \textbf{Answer Template: } \{object\} \\ \midrule
\textbf{Mapping Templates:} \\ 
\{subject\} is employed in \{object\}\\ 
\{subject\} earns a living by working as \{object\}\\ 
\{subject\}'s occupation is \{object\}\\ 
\{subject\} is engaged in \{object\}\\ 
\{subject\}'s profession is \{object\}\\ 
\{subject\} works as a \{object\}\\ 
\{subject\} makes a living as \{object\}\\ 
\{subject\} has a career in \{object\}\\ 
\{subject\} is involved in \{object\}\\ 
\{subject\} engages in the occupation of \{object\}\\ 
The work that \{subject\} undertakes is classified as \{object\}\\ 
The focus of \{subject\}'s employment lies in \{object\}\\ 
The type of work \{subject\} engages in is \{object\}\\ 
The work performed by \{subject\} falls under \{object\}\\ 
The work done by \{subject\} falls under the category of \{object\}\\ 
The kind of work \{subject\} does is \{object\}\\ 
\{subject\} operates in the field of \{object\}\\ 
The work \{subject\} performs is \{object\}\\ 
When it comes to work, \{subject\} does \{object\}\\ 
\{subject\} works in the field of \{object\}\\ 
The work done by \{subject\} is \{object\} \\ \bottomrule
\end{tabular}
\label{tab:P106}
\end{table}

\begin{table}[H]
\centering
\caption{Information and mapping templates for topic P264 (out-of-domain).}
\begin{tabular}{p{0.95\linewidth}}
\toprule
\textbf{Topic:} P264 \\ \textbf{Question Template: } What music label is \{subject\} represented by? \\ \textbf{Answer Template: } \{object\} \\ \midrule
\textbf{Mapping Templates:} \\ 
\{subject\} is represented by \{object\}\\ 
The music label representing \{subject\} is \{object\}\\ 
Regarding representation, \{subject\} is under \{object\}\\ 
\{subject\} has a record deal with \{object\}\\ 
\{subject\} has a musical partnership with \{object\}\\ 
\{subject\}'s music is released by \{object\}\\ 
\{subject\} is signed to \{object\}\\ 
\{subject\} is affiliated with \{object\}\\ 
\{subject\} has a contract with \{object\}\\ 
\{subject\} is represented by the music label \{object\}\\ 
The talented \{subject\} is associated with the music label \{object\}\\ 
\{subject\}'s discography is managed by the renowned label \{object\}\\ 
\{subject\} is under contract with the music label \{object\}\\ 
\{subject\} is affiliated with the music label \{object\}\\ 
The music label backing \{subject\} is \{object\}\\ 
\{subject\} is signed with the music label \{object\}\\ 
\{subject\} works with the music label \{object\}\\ 
\{subject\} is under the music label \{object\}\\ 
The music label that represents \{subject\} is \{object\}\\ 
\{subject\} has representation from \{object\}\\ 
If you were to ask what music label represents \{subject\}, it is \{object\} \\ \bottomrule
\end{tabular}
\label{tab:P264}
\end{table}

\begin{table}[H]
\centering
\caption{Information and mapping templates for topic P407 (out-of-domain).}
\begin{tabular}{p{0.95\linewidth}}
\toprule
\textbf{Topic:} P407 \\ \textbf{Question Template: } Which language was \{subject\} written in? \\ \textbf{Answer Template: } \{object\} \\ \midrule
\textbf{Mapping Templates:} \\ 
\{subject\} was originally written in \{object\}\\ 
The language used for writing \{subject\} was \{object\}\\ 
The original text of \{subject\} appeared in \{object\}\\ 
\{subject\} was penned in \{object\}\\ 
The language of \{subject\} is \{object\}\\ 
\{subject\} was composed in \{object\}\\ 
\{subject\} was created in \{object\}\\ 
\{subject\} is written in the language of \{object\}\\ 
The writing language of \{subject\} is \{object\}\\ 
\{subject\} was composed in the language known as \{object\}\\ 
The linguistic medium of \{subject\} is \{object\}\\ 
The choice of language for \{subject\} is \{object\}\\ 
\{subject\} was written in the language of \{object\}\\ 
The language used to write \{subject\} is \{object\}\\ 
The original language of \{subject\} is \{object\}\\ 
The writing of \{subject\} is in \{object\}\\ 
\{subject\} is composed in \{object\}\\ 
The text of \{subject\} is in \{object\}\\ 
\{subject\} was written in \{object\}\\ 
If you were to ask what language \{subject\} was written in, it's \{object\}\\ 
Looking at the language of \{subject\}, it's \{object\} \\ \bottomrule
\end{tabular}
\label{tab:P407}
\end{table}

\begin{table}[H]
\centering
\caption{Information and mapping templates for topic P800 (out-of-domain).}
\begin{tabular}{p{0.95\linewidth}}
\toprule
\textbf{Topic:} P800 \\ \textbf{Question Template: } What is \{subject\} famous for? \\ \textbf{Answer Template: } \{object\} \\ \midrule
\textbf{Mapping Templates:} \\ 
\{subject\} is famous for \{object\}\\ 
The fame of \{subject\} is due to \{object\}\\ 
People recognize \{subject\} for \{object\}\\ 
\{subject\} is renowned for \{object\}\\ 
\{subject\}'s claim to fame is \{object\}\\ 
\{subject\} is celebrated for \{object\}\\ 
\{subject\} is known for \{object\}\\ 
\{subject\} is distinguished by \{object\}\\ 
\{subject\} is admired for \{object\}\\ 
Fame comes to \{subject\} due to \{object\}\\ 
Among its achievements, \{subject\} is celebrated for \{object\}\\ 
\{subject\}'s popularity largely stems from \{object\}\\ 
\{subject\}'s notable recognition comes from \{object\}\\ 
\{subject\} is celebrated widely due to \{object\}\\ 
The fame of \{subject\} is attributed to \{object\}\\ 
The reason \{subject\} is famous is \{object\}\\ 
\{subject\} is well-known for \{object\}\\ 
\{subject\} gained fame for \{object\}\\ 
If you were to ask what \{subject\} is famous for, it's \{object\}\\ 
Looking at what made \{subject\} famous, it's \{object\}\\ 
In terms of fame, \{subject\} is associated with \{object\} \\ \bottomrule
\end{tabular}
\label{tab:P800}
\end{table}

\begin{table}[H]
    \centering
    \caption{The template for SFT of each LLM. `\{Question\}' and `\{Answer\}' represent to the question and answer of each sample, respectively.}
    \begin{tabular}{p{0.17\linewidth}m{0.78\linewidth}}
    \toprule
    \textbf{Models} & \textbf{Templates} \\ \midrule
        {LLaMA-2-7B} & <s>[INST] \{Question\} [/INST] \{Answer\} </s> \\\midrule
        {LLaMA-2-13B} & <s>[INST] \{Question\} [/INST] \{Answer\} </s> \\ \midrule
        {LLaMA-3-8B} & <|begin\_of\_text|><|start\_header\_id|>user<end\_header\_id|>\newline \{Question\}<|eot\_id|><|start\_header\_id|>assistant<|end\_header\_id|>\newline \{Answer\}<|eot\_id|><|end\_of\_text|>\\ \midrule
        {Qwen-2-7B} & <|im\_start|>user\newline \{Question\}<|im\_end|>\newline <|im\_start|>assistant\newline \{Answer\}<|im\_end|>\\ \bottomrule
    \end{tabular}
    \label{tab:prompt}
\end{table}

\end{document}